\documentclass[10pt,journal,compsoc]{IEEEtran}
\ifCLASSOPTIONcompsoc
  \usepackage[nocompress]{cite}
\else
  \usepackage{cite}
\fi
\ifCLASSINFOpdf
  \usepackage[pdftex]{graphicx}
\else
  \usepackage[dvips]{graphicx}
\fi
\usepackage{amsmath}
\usepackage{xcolor}
\usepackage{soul}
\usepackage{subfig}

\begin{document}
%

\title{Improving Recurrent Neural Network Responsiveness to Acute Clinical Events}

%
%
%

\author{David~Ledbetter, Eugene~Laksana, Melissa~Aczon,
        and~Randall~Wetzel
\IEEEcompsocitemizethanks{\IEEEcompsocthanksitem All four authors are with the Laura P. and Leland K. Whittier Virtual Pediatric Intensive Care Unit, Children's Hospital Los Angeles
CA, 90027. They can be reached at: \protect\\
 \{dledbetter, elaksana, maczon, rwetzel\}@chla.usc.edu 
}
\thanks{}}

\IEEEtitleabstractindextext{%
\begin{abstract}
Predictive models in acute care settings must be able to immediately recognize precipitous changes in a patient's status when presented with data reflecting such changes. Recurrent neural networks (RNN) have become common for training and deploying clinical decision support models. They frequently exhibit a delayed response to acute events. New information must propagate through the RNN's cell state memory before the total impact is reflected in the model's predictions. This work presents input data perseveration as a method of training and deploying an RNN model to make its predictions more responsive to newly acquired information: input data is replicated during training and deployment. Each replication of the data input impacts the cell state and output of the RNN, but only the output at the final replication is maintained and broadcast as the prediction for evaluation and deployment purposes. When presented with data reflecting acute events, a model trained and deployed with input perseveration responds with more pronounced immediate changes in predictions and maintains globally robust performance. Such a characteristic is crucial in predictive models for an intensive care unit.  
\end{abstract}

\begin{IEEEkeywords}
Recurrent Neural Network, Long Short-Term Memory, Acute Clinical Events, Electronic Medical Records.
\end{IEEEkeywords}}

\maketitle

\IEEEdisplaynontitleabstractindextext

%
\IEEEpeerreviewmaketitle

\IEEEraisesectionheading{\section{Introduction}\label{sec:intro}}

\subsection{The Problem}

Critical care environments require rapid decision making. To be meaningful, predictive models in these settings must immediately recognize precipitous changes in a patient's state when presented with data reflecting such changes  \cite{chan2008hospital,jones2011rapid}. Newly acquired information must be integrated quickly with the existing understanding of the patient's state to inform clinical decisions. 

Long Short-Term Memory (LSTM) models \cite{hochreiter1997long} are a type of Recurrent Neural Network (RNN) architecture and have become increasingly popular for modeling tasks in clinical settings, including Intensive Care Units (ICUs) \cite{rajkomar2018scalable,esteva2019guide,oh2018automated,faust2018deep,laksana2019impact,miotto2017deep,aczon2017dynamic,carlin2018predicting}. It is occasionally noted that RNNs, including LSTM models, exhibit some lag after acquiring new information \cite{flovik_2018}. This characteristic is frequently observed in stock prediction tasks where the apparent predictive capability of a model is comparable to a model that predicts the last observed stock price \cite{kim2018forecasting,xiong2015deep}. However, the intersection of deep learning and healthcare, though ever widening, has a paucity of literature on this phenomenon. Commonly reported metrics such as Area Under the Receiver Operating Characteristic Curve, sensitivity, specificity, and mean absolute errors mask certain behaviors of RNN based models.

RNN models require computational cycles to incorporate new data into their memory cell \cite{olah2015understanding}. There are two primary observable effects of this, summarized briefly here. First, what we refer to as a \textit{pipe-up} period, requires the initial input to propagate through the memory cell to overcome the initialized cell state before that input significantly influences the predictions. Second, during all subsequent times, new information only slightly changes the overall prediction of a model at the time of input and requires propagation through the memory cell before fully augmenting the model's prediction. The first phenomenon is a special case of the second. These two effects are further illustrated in Sections \ref{sec:assess} and \ref{sec:results}.

\subsection{Proposed Solution}
This work aims to enable more pronounced changes in RNN predictions when the model is presented with data indicating acute clinical events by \emph{perseverating} the input. Instead of giving newly available input data to an RNN model only once, the input is replicated and given to the model multiple times, during both training and deployment, with only the prediction of the final perseveration maintained and made visible to the end-user. This approach provides additional computational cycles for new data to be incorporated into the model's internal states before a prediction is broadcast. We hypothesized that the resultant model would be able to react to acute events more quickly than traditionally trained LSTMs and still maintain overall model performance. 

%
%
%
%


\section{Related Works}
\label{sec:related}

Since the seminal paper by R.E. Kalman in 1960 describing what is now called the Kalman Filter \cite{kalman1960new}, engineers have been aware of the trade off between integrating historical data and responding to new data. Most training techniques used in modern deep learning affect the balance between relying on historical trends and responding to new information; such techniques include dropout, optimizers, and activation functions  \cite{srivastava2014dropout,tieleman2012lecture,kingma2014adam,nair2010rectified}. Generating an appropriate target vector (e.g. using changes in values between consecutive time points instead of predicting raw values) is another training technique that can prevent generation of an auto-correlation model \cite{flovik_2018}.

Attention mechanisms \cite{luong2015effective} are sometimes used to manage the balance between historical and new data by appending the entire input sequence to the final hidden layer of an RNN. Doing so affords the model another opportunity to learn weights which expose the moments in time series data most relevant to the predictions. Attention networks were originally developed for post-hoc sequence-to-sequence modeling tasks such as image captioning and neural machine translation which permit access to the entire input sequence when making predictions \cite{xu2015show,luong2015effective}. An attention mechanism was applied by Zhang, et al. to predict risk of future hospitalization using medical records from a fixed observation window.\cite{zhang2018patient2vec}. 
\section{Data}
\subsection{Clinical Data Sources}
Data for this work were extracted from de-identified observational clinical data collected in Electronic Medical Records (EMR, Cerner) in the Pediatric Intensive Care Unit (PICU) of Children's Hospital Los Angeles (CHLA) between January 2009 and February 2019. The CHLA Institutional Review Board (IRB) reviewed the study protocol and waived the need for IRB approval.  A patient record included static information such as gender, race, and discharge disposition at the end of an ICU \emph{episode},  defined as a contiguous admission in the PICU. A patient may have multiple episodes. The EMR for an episode also contained irregularly, sparsely and asynchronously \textit{charted} measurements of physiologic observations (e.g. heart rate, blood pressure), laboratory results (e.g. creatinine, glucose level), drugs (e.g. epinephrine, furosemide) and interventions (e.g. intubation, oxygen level).  Episodes without discharge disposition were excluded, leaving 12,826 episodes (9,250 patients) in the final dataset. 

Prior to any of the computational experiments described here, the episodes were randomly partitioned into three sets: a training set for deriving model weights, a validation set for optimizing hyper-parameters, and a holdout test set for measuring performance. To minimize bias in the performance evaluation metrics, partitioning was done at the patient level, i.e. all episodes from a single patient belonged to only one of these sets: 60\% in the training set, 20\% in the validation set, and 20\% in the test set. No other stratification was applied. Table \ref{tab:demo} displays basic characteristics of the resulting data partitions.

Preprocessing steps described in previous work \cite{ho2017dependence} converted each episode data to a matrix structure amenable to machine learning model development. A row of values in this matrix represented measurements (recorded or imputed) of different variables at one particular time, while a column contained values of a single variable at different times. A complete list of variables used as model inputs can be found in Appendix \ref{appendix:EMR_vars}. Note that diagnoses, while available in the EMR, were not used as input features.

\begin{table}[!t]
\caption{Basic demographics of data used in this study for each data partition.}
\label{tab:demo}
\centering
\begin{tabular}{|l|lll|l|} \hline
    \textbf{}          & \textbf{training} & \textbf{validation} & \textbf{test} & \textbf{overall} \\ \hline
    Number of episodes & 7663           & 2541           & 26221           & 12826 \\
    Number of patients & 5551           & 1850           & 18501           & 9251 \\
    Mortality rate     & 0.035          & 0.037         & 0.043            & 0.037 \\
    Fraction male      & 0.56           & 0.55           & 0.58           & 0.56          \\
    Age mean (years) & 7.99 & 8.23 & 8.09 & 8.06   \\
    \hspace{0.5cm} std dev & 6.45 & 6.51 & 6.35 & 6.44 \\\hline
    \end{tabular}
\end{table}


\subsection{Target Outcome}

ICU mortality was chosen as the target outcome because it is a simple, unambiguous outcome. Importantly, 
risk of mortality is used commonly as a proxy for severity of illness in critical care
\cite{pollack1996prism,pollack2016pediatric, slater2003pim2, leteurtre2015daily}.
The overall mortality rate of the data was $3.7\%$ (Table \ref{tab:demo}). 


\section{RNN Models}

Many-to-many recurrent neural network models, consisting of stacked Long Short-Term Memory (LSTM) layers followed by a dense layer, were trained to predict ICU mortality of each patient episode. All the models output a probability of survival at each distinct time point where an observation or measurement of the patient was made, generating a trajectory of scores that reflect the changing status of a patient during their ICU episode.

\begin{figure}[!t]
\centering
    \includegraphics[width=3.4in]{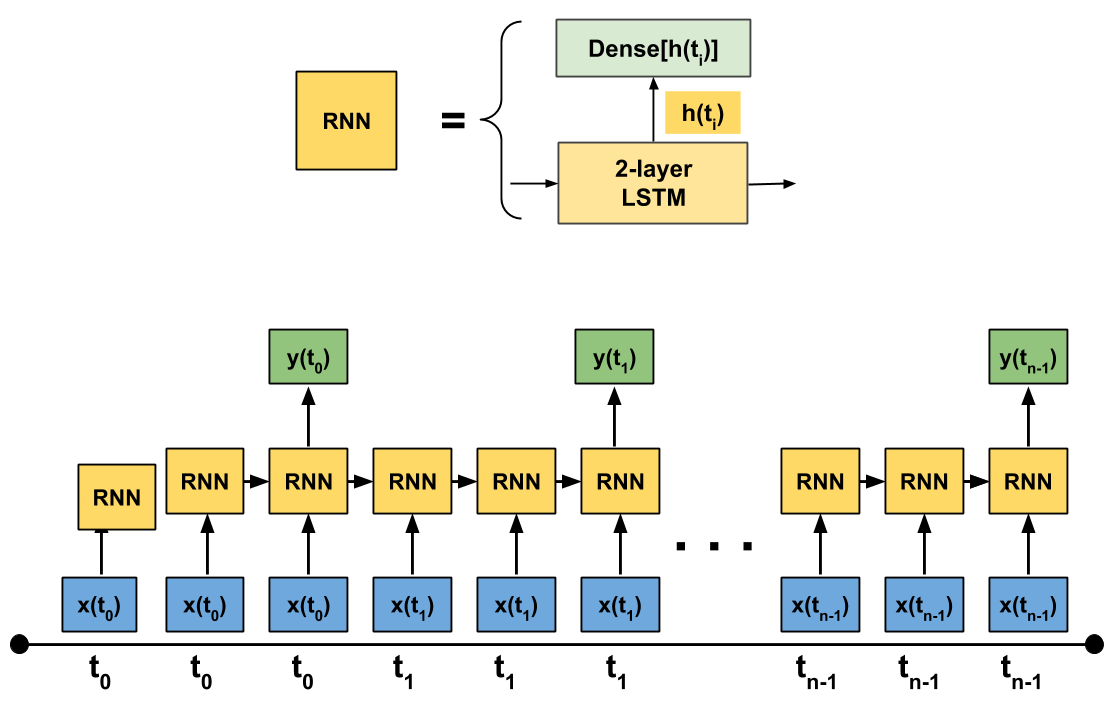}
    \caption{The perseverating recurrent neural network (PRNN) is an RNN with repeating inputs.  Each input vector associated with a particular time is replicated $k$ times, and only the output from the last replication is considered as the prediction for that time.  The diagram above illustrates the process for $k=3$.} 
    \label{fig:data}
\end{figure}

\begin{figure}[!t]
\centering
    \includegraphics[width=1.8in]{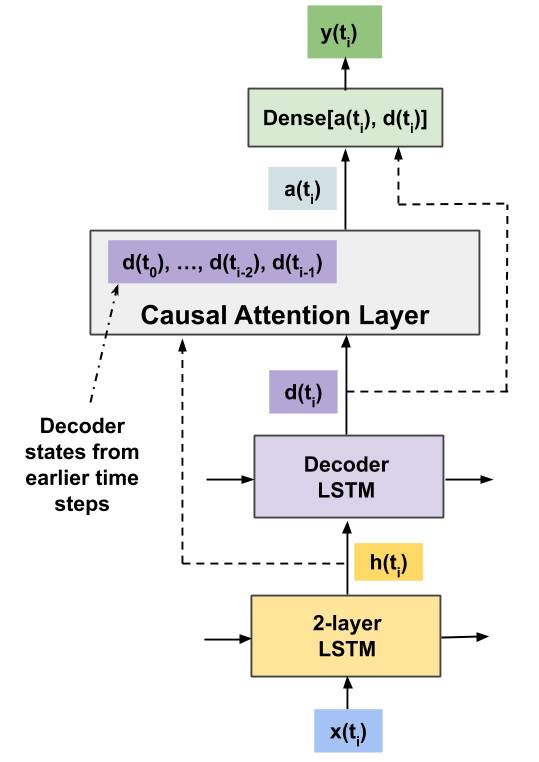}
    \caption{Overview of a standard RNN model with a causal attention layer.} 
    \label{fig:attention_model}
\end{figure}

The \textit{baseline} RNN model was trained in the traditional manner: when the model acquires new data $x(t_i)$, it generates and immediately broadcasts a prediction $y(t_i)$. 
Other models which share the architecture of the baseline RNN model were trained using input \textit{perseveration}: 
the same $x(t_i)$ vector was repeatedly given as input to the RNN model $k$ times, where $k$ is a controllable parameter. 
While all $k$ outputs -- corresponding to the $k$ times that $x(t_i)$ was given to the model -- are used for optimization during training, only the last one is maintained and broadcast as the prediction at time $t_i$ during performance assessment and deployment. We call these models perseverating recurrent neural networks (PRNN). Figure \ref{fig:data} illustrates a PRNN model with $k=3$. Note that the baseline RNN can be considered as a PRNN with $k=1$.
Input perseveration provides the internal cell state memory of the RNN additional computational cycles to incorporate the current state of the patient into the final prediction.

An attention network using the same hyperparameters of the baseline RNN model was also implemented. Most attention networks have access to the entire sequence of inputs -- both past and future relative to the current one -- when making a prediction at any point in time.  
However, continuous monitoring of patient status precludes access to future information, available in a retrospective study but not in a real deployment scenario. Therefore, the  
attention layer between the last hidden layer and output layer used a \emph{causal mask} that exposed only the inputs up to the time when a prediction is being made \cite{vaswani2017attention}. This ensured that the network did not use future information to make its predictions while affording it another opportunity to consider the totality of information up to the current time.
See Figure \ref{fig:attention_model} for a diagram of the model. 


All six models -- baseline RNN ($k=1$), PRNN for $k=2,3,4,5$, and attention RNN -- were implemented and trained using Keras 2.0.7 with the Theano 1.0.2 backend \cite{keras2018, 2016arXiv160502688short}. For each model, weights were optimized on the training and validation sets; performance was computed on the validation set after every epoch (i.e. a full cycle through the training set), and the best performing weights were saved as the final model. Table \ref{tab:hyperparam} displays the hyperparameters that were used for all six models. The final models were assessed for performance metrics on the test set.

\begin{table}[!t]
\caption{Hyperparameters for all permutations of the PRNN.}
\label{tab:hyperparam}
\centering
    \begin{tabular}{|l|l|}
    \hline
    \textbf{Hyperparameter}     & \textbf{Value}       \\
    \hline                                             
    Number of LSTM Layers       & 3                    \\
    Hidden Units in LSTM Layers & {[}128, 256, 128{]}  \\
    Batch Size                  & 32                   \\
    Initial Learning Rate               & 1e-5                 \\
    Loss                        & binary cross-entropy \\
    Optimizer                   & rmsprop              \\
    Dropout                     & 0.2                  \\
    Recurrent Dropout           & 0.2                  \\
    Regularizer                 & 0.0001               \\
    Output Activation           & sigmoid              \\       
    \hline
    \end{tabular}
\end{table}

\section{Model Assessments}
\label{sec:assess}

Standard metrics such as the Area Under the Receiver Operating Characteristic Curve (AUROC), precision, and recall scores for a binary classification task can capture a model's overall predictive performance. However, they can mask the lag phenomenon displayed by LSTMs that, although known, is rarely commented on in the literature. 
The trajectory of probability of survival predictions should reflect the evolving state of a patient which can have instantaneous changes. Metrics were designed to quantify a model's pipe-up behavior and its ability to capture rapid changes resulting from clinically adverse events. 
AUROC was used to compare overall predictive performance for the main task -- ICU mortality prediction.

\subsection{Pipe-Up Behavior}

A model's prediction at the first time step is a function of its weights, initial state of its memory cells and the first input. Of these, only the first input varies across different patient episodes; therefore, the distribution of a model's prediction at the first time point of all episodes indicates the model's level of reliance on the first input data. We refer to this period as the \textit{pipe-up} period. 

The mean and standard deviation of the distribution of all $\hat{y}_p(t_0)$ predictions, where $\hat{y}_p(t_0)$ represents the first prediction for patient episode $p$, were computed for all survivors and non-survivors in the test set. These metrics were computed for all six RNN models described in the previous section and used to compare their responsiveness to the first available information about a patient episode.

\subsection{Responsiveness to Acute Events}\label{sec:variation_metric}

The models were also assessed for their instantaneous responses to clinically adverse events using an average variation metric computed from the predictions during such periods.
%
%
In consultation with clinicians, the times when any of the following occurred were defined as acute time points or intervals:
\begin{enumerate}
    \item \textit{substantial} increases in creatinine levels or inotrope score between two consecutive recordings; 
    \item \textit{substantial} decreases in heart rate, mean arterial pressure, Glasgow Coma Score, blood oxygen level (SpO2), arterial blood gas (ABG) pH, and venous blood gas (VBG) pH between two consecutive recordings;
    \item a patient's heart rate goes to 0.
\end{enumerate}

We quantified \textit{substantial} changes as those that were in the top or bottom $X$ percentile ($X \in \{5.0, 1.0, 0.5\}$) of inter-measurement changes.
For example, a time point $t_i$ was considered acute in terms of creatinine when the \textit{increase} in creatinine level from $t_{i-1}$ to $t_i$ was in the top $X$ percentile of all creatinine level percent changes between any two consecutive time points in the test dataset.
Similarly, acute time points for heart rate were those times when the \textit{decrease} in heart rate was in the bottom $X$\% of all heart rate changes between any two consecutive time points of all test set episodes. Table \ref{tab:thresholds} shows the specific percent change thresholds indicating \emph{acute} changes in a patient. Thus, a 32\% drop in heart rate between two consecutive measurements would be in the top 1 percentile, while a 70\% increase in inotrope score would be in the top 5 percentile.

Applying three percentile-based thresholds to define acute patient state changes meant a total of 25 acuity definitions (2 * 3 increases in measured variables, 6 * 3 decreases in measured variables, and zero heart rate) used to assess models' responses when such events occur. It is important to note that we are not proposing these as general definitions of clinical acuity. They were designed to capture events that indicate precipitous changes in a patient's state with high specificity to assess the magnitude of the instantaneous change in predicted probability of survival.

Each previously defined acute event, denoted by $E$, identified a set of \emph{acute time points}, denoted by $S_E(p)$, for each patient episode $p$. Changes in a model's prediction at these time points were evaluated through an average temporal variation metric given by:
\begin{eqnarray}
    V_E(p) &=& \frac{100}{|S_E(p)| }\sum_{t_i \in S_E(p)} |\hat{y}_p(t_i) - \hat{y}_p(t_{i-1})|,\label{eq:var_metric}
\end{eqnarray}
where $\hat{y}_p(t_i)$ is the prediction for patient episode $p$ at time point $t_i$, and $|S_E(p)|$ is the number of time points in $S_E(p)$. 
This metric measures how much the predicted probability of survival (scaled to $[0,100]$) changed, on average, at the defined acute time points of an episode. Model $A$ having a higher $AV(p)$ than model $B$ means that when both models were presented with data reflecting a precipitous change in a patient's state, model $A$'s prediction underwent a greater change than model $B$'s prediction, indicating that model $A$ had a more pronounced response to the acute event. The average of this metric across $N_p$ episodes gives a measure of the model's overall responsiveness to acute event $E$ of those episodes:
\begin{equation}
    \bar{V}_E = \frac{1}{N_p}\sum_{p} V_E(p). \label{eq:var_metric_all}
\end{equation}



\begin{table}[!t]
\caption{Thresholds of percent change used to define acute changes in patient state. Negative thresholds correspond to the bottom $p\%$ of changes and positive thresholds correspond to the top $p\%$  of changes, for $p\in[5, 1, 0.5]$.}
\label{tab:thresholds}
\centering
\begin{tabular}{|l|lll|} \hline
                           & 5\%     & 1\%     & 0.5\%  \\ \hline
    Heart Rate            & -18.8 & -31.6 & -37.1 \\
    Mean Arterial Pressure & -23.5 & -38.1 & -44.2 \\
    Glascow Coma Score     & -50.0  & -72.7 & -72.7 \\
    Pulse Oximetry         & -5.6  & -20.2 & -35.1 \\
    ABG pH                 & -1.5  & -2.9  & -3.5  \\
    VBG pH                 & -1.4  & -2.6  & -3.1  \\
    Inotrope Score        & 66.7   & 150.0  & 250.0  \\
    Creatinine             & 33.3  & 60.5  & 81.8  \\ \hline
\end{tabular}
\end{table}

\subsection{Overall Predictive Performance}

Model predictions at ICU admission and at 1, 3, 6, 12, and 24 hours after ICU admission were evaluated for predictive performance via the AUROC. Performance was only measured for episodes lasting at least $24$ hours to maintain a consistent cohort across time slices ($n=2130$, mortality rate = 0.046).

\section{Results}
\label{sec:results}

\begin{figure}[ht]
    \centering
    \includegraphics[width=\linewidth]{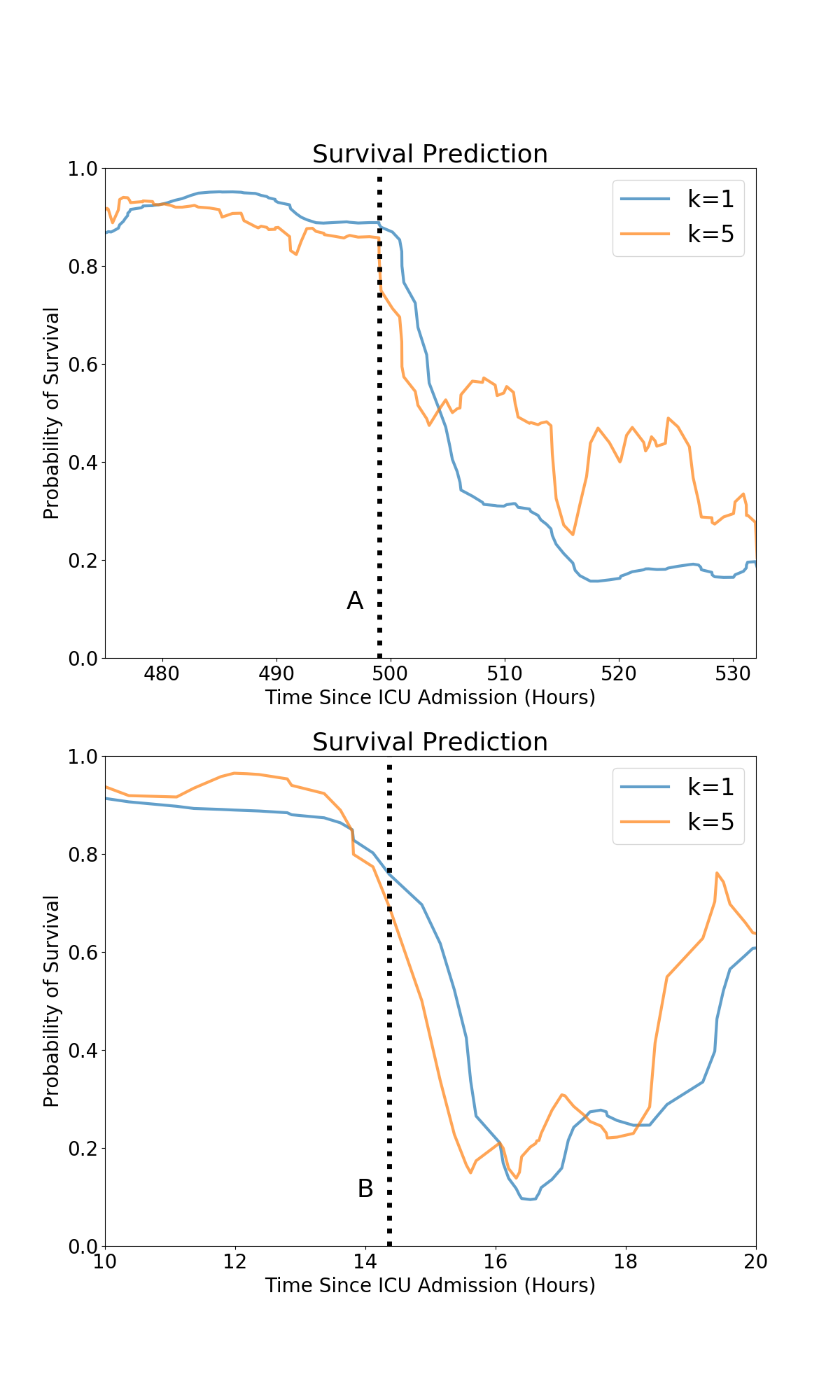}
    \caption{Two examples of predictions following acute events in two individual patient episodes. The blue curves correspond to the baseline ($k=1$) model, while the orange curves correspond to the PRNN model with $k=5$. The dotted vertical lines denote onset of an acute event for each patient: (A) a patient experiencing kidney failure and the moment of increased creatinine and blood urea nitrogen (BUN); (B) a patient's Glasgow Coma Score decreasing from 7 to 4, indicating significantly reduced levels of consciousness.\vspace{0.6cm}
    \label{fig:examples}}
\end{figure}

Figure \ref{fig:examples} illustrates trajectories of probability of survival predictions from two models, the baseline RNN ($k=1$) and a PRNN with $k=5$, for two non-surviving episodes. In the first episode (top), the patient had a rapid increase of creatinine level, which indicates kidney dysfunction, about 500 hours after ICU admission. When presented with this defined acute event, the baseline model's prediction dropped from 0.89 to 0.88, while the PRNN model dropped from 0.75 to 0.71.
In the second episode (bottom), the patient's Glasgow Coma Score decreased from 7 to 4. When presented with this defined acute event, the baseline model's probability of survival prediction dropped from 0.76 to 0.70, while the PRNN model dropped from 0.69 to 0.50.
In either case, the PRNN model's immediate response to an acute event was more pronounced than that of the baseline RNN model. Further, the trajectory of the baseline RNN's predictions after either acute event appears to lag behind that of the PRNN by about 1-2 hours.

The remainder of this section describes the results from aggregating the assessment metrics described in Section \ref{sec:assess} across episodes in the test dataset.

\begin{figure*}
    \centering
    \includegraphics[width=\linewidth]{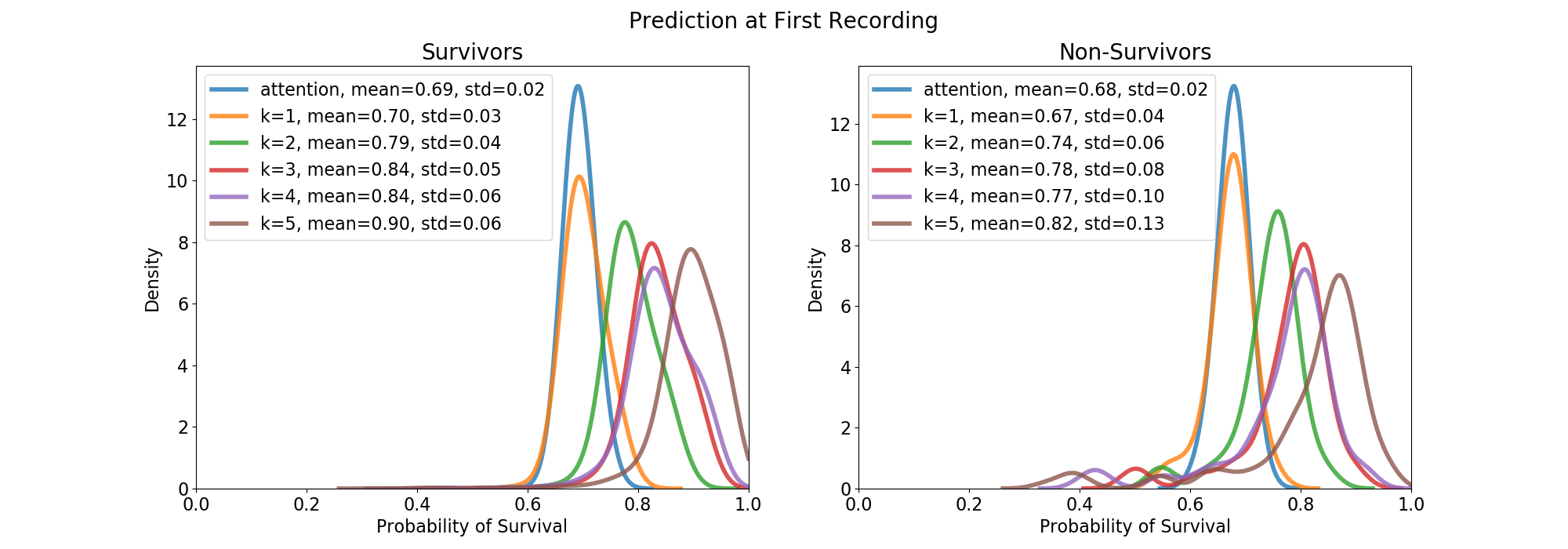}
    \caption{Distribution of model predictions at the first available observation for survivors and non-survivors} \vspace{0.6cm}
    \label{fig:t0_dists}
\end{figure*}

\subsection{Pipe-Up Behavior}

Figure \ref{fig:t0_dists} shows the distribution of predictions at the first observation. The mean prediction for survivors increased with the perseveration parameter $k$: from $0.70$ ($k=1$) to $0.90$ ($k=5$). For both survivors and non-survivors, the standard deviation increased with $k$. The increase of standard deviation was more pronounced in the non-survivors, going from from $0.04$ at $k=1$ to $0.13$ at $k=5$.

\subsection{Responsiveness to Acute Events}
\begin{figure*}
    \centering
    \includegraphics[width=\linewidth]{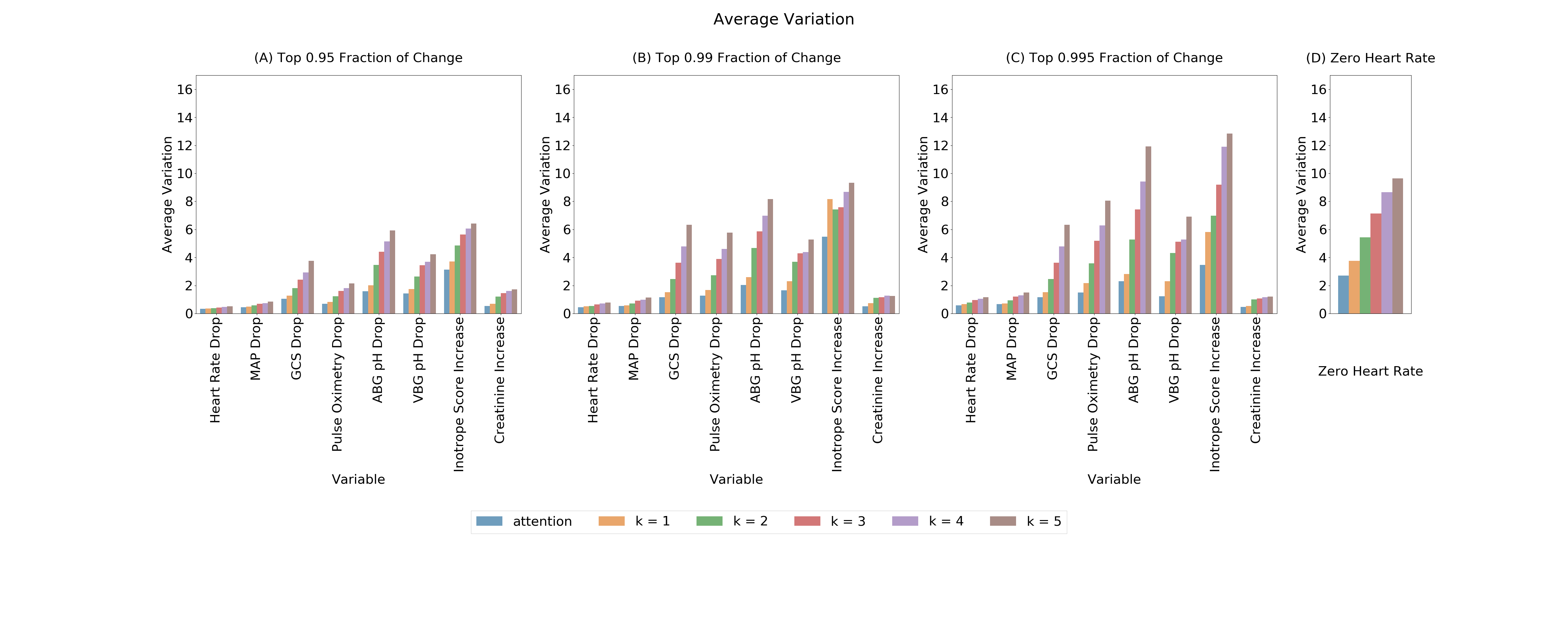}
    \caption{Comparison of the average variation metric $\bar{V}_E$ of different models as defined by Equation \ref{eq:var_metric_all} across all test set episodes, where the events are defined by: A) a 95th percentile change; B) a 99th percentile change; C) a 99.5th percentile change; D) a cardiac arrest.}
    \vspace{0.6cm}
    \label{fig:variation}
\end{figure*}

Figure \ref{fig:variation} displays the average variation as a function of $k$ and acuity definitions.  Two trends are apparent. First is the behavior of average variation as a function of percentile change in a given physiologic observation or intervention. For each model, the resulting average variation increased as the change in a physiologic or intervention variable became more severe (ie. from 95.0 to 99.5 percentile). Second, a clear positive correlation between $k$ and average variation is evident across all definitions of acuity.  The attention layer generally demonstrated lower average variation than the $k=1$ model. 

\subsection{Overall Predictive Performance}
Table \ref{tab:aucs} summarizes the AUROC of model predictions at different times after ICU admission of all episodes in the test set that lasted at least 24 hours. The AUROC gains due to input perseveration were larger at the earlier times of prediction ($t \leq 6$ hours), increasing at the first hour from $0.77$ when $k=1$ to $0.83$ when $k=5$. By the 12th hour, the AUROC did not significantly change with the perseveration parameter $k$. Performance increases for all models as additional hours of observation are available. Adding an attention layer displayed similar performance as the baseline model ($k=1$) at all evaluation times.


\begin{table}[]
\caption{Test set AUROCs for the mortality prediction at admission and after $1$, $3$, $6$, $12$, and $24$ hours of observation. Only the patient episodes lasting at least $24$ hours of ICU time were evaluated.}
\label{tab:aucs}
\centering
\begin{tabular}{|l|llllll|}
\hline
\textbf{k} & \textbf{0hr} & \textbf{1hr} & \textbf{3hr} & \textbf{6hr} & \textbf{12hr} & \textbf{24hr} \\ \hline
1          & 0.716 & 0.766       & 0.859       & 0.895       & 0.914        & 0.952        \\
2          & 0.729 & 0.804       & 0.871       & 0.902       & 0.916        & \textbf{0.956}        \\
3          & 0.731 & 0.819       & 0.875       & 0.902       & 0.916        & 0.955        \\
4          & 0.735 & 0.821       & 0.876       & 0.900       & 0.915        & 0.954        \\
5          & 0.736 & \textbf{0.832}       & \textbf{0.883}       & \textbf{0.905}       & \textbf{0.919}        & 0.955 \\  \hline \hline  
attention  & 0.723 & 0.734             & 0.853             & 0.892             & 0.910              & 0.949  \\ \hline 
\end{tabular}

\end{table}
\section{Discussion}
Real time predictions for ICU mortality are proxies for severity of illness \cite{leteurtre2015daily, badawi2018evaluation, hug2009icu} and should reflect acute changes in a patient. The examples in Figure 3 show that the mortality model trained with input perseveration (PRNN with $k=5$) responded more pronouncedly and immediately than the traditionally trained model ($k=1$) when both were presented with data reflecting acute changes. Subsequent to the acute changes, the responses of the traditional RNN appeared to lag behind the PRNN's. Standard metrics such as AUROCs for classification or mean absolute errors for continuous regression often mask this predictive lag and other deleterious behaviors that can be detrimental in critical or intensive care settings where rapid recognition and response are crucial. 

Input perseveration provides the RNN's internal cells additional computational cycles to incorporate the current state of the patient into the final prediction. 
The effect of this method was measured by metrics designed to capture a model's immediate responses to newly acquired data.

The variation metric described in Section \ref{sec:variation_metric} compares the models' immediate responses to data indicating precipitous changes in a patient's status. These changes require quick reaction time from the care team, therefore capturing them in the predictions is important. The comparison of variation metrics from the different models (Figure \ref{fig:variation}) shows that the LSTM became more responsive to acute clinical events as the level of perseveration, $k$, increased. When $k$ increased from 1 to 5, the variation metric corresponding to the defined precipitous events increased by a factor of 2-3 times. They also show that a given model's responsiveness increased with more acute events (i.e. those belonging to higher percentile changes for a given physiologic or intervention variable). This result is consistent with expectations about the variation metric. For example, one would expect the change in predictions of those with larger drops in blood gas level to be greater than those of less severe drops. 

Figure \ref{fig:t0_dists} compares the distributions of initial predictions generated by the different models. Since the first prediction is a function of both the initial input (which varies across episodes) and the initial cell state memory (which is fixed), a wider distribution of these predictions across episodes indicates a higher reliance on the initial input. This is important because children admitted to the PICU have different severities of illness \cite{pollack1996prism, slater2003pim2}. Increasing $k$ resulted in a wider distribution as measured by the standard deviation. The increase was greater for the non-surviving population ($\sigma=0.04$ when $k=1$ to $\sigma=0.13$ when $k=5$) relative to the surviving population. This is again consistent with clinical expectations.  

Increasing the first prediction's reliance on the initial measurement -- as achieved by the PRNN models with higher $k$ -- also resulted in higher AUROC at ICU admission (Table \ref{tab:aucs}). As the models observed the patient longer, their AUROCs increased as expected. Importantly, the increase of AUROC due to perseveration was greater at the early hours ($t\leq3$) when information is most scarce, with the 1-hour AUROC increasing from 0.77 when $k=1$ to 0.83 when $k=5$. This means that predictive clinical models relying on scarce data for early detection could benefit from the perseveration approach.

Adding a causal attention layer to the baseline ($k=1$) model had no apparent performance improvement in any of the metrics. The attention network theoretically can put more weight to the most recent state than to the previous ones. The results indicate that this mechanism did not improve on what the baseline LSTM's gates were already doing, but persistently giving the same input to the model -- i.e. perseverating the input  -- did.

There are limitations to the perseverated input approach. Although LSTMs are theoretically good at understanding temporal trends through their memory cells, there remain practical limitations to how long prior information can be maintained to inform future predictions \cite{zhang2016highway,yu2017long}. The PRNN has the potential to exacerbate these algorithmic deficiencies because of the memory cell's prolonged exposure to the same data. A comprehensive assessment of the impact of a potential reduction in temporal memory was not performed. However, the basic data perseveration technique can easily be generalized to any sequential data.

Additionally, perseveration increases the number of sequences requiring computation. Perseveration is used during both training and inference, and the compute time scales linearly for both, commensurate with the level of perseveration. However, the time required for inference is on the order of $30ms$ for $k=1$ (baseline) and $150ms$ for $k=5$ on an NVIDIA Titan RTX. Since the recording frequency is approximately every 15 minutes, these computational burdens do not hinder deployment. 

Finally, this proof-of-concept study used only a single clinical outcome (ICU mortality) and data from a single center. Future work will examine the effect of perseverating the data input on other important clinical tasks such as risk of desaturation, sepsis, and renal failure.

\section{Conclusion}

This work demonstrates that perseverated data input increases the responsiveness of LSTM models to a variety of acute changes to patient state and also significantly increases predictive performance in the early hours following admission. The PRNN is a simple solution to the predictive lag exhibited by standard LSTMs when encountering an acute event and may enable more rapid responses to critical conditions.  

\label{sec:conclusion}


\section*{Acknowledgements}

This work was supported by the L. K. Whittier Foundation. The authors would like to thank Alysia Flynn for help aggregating the data and Mose Wintner for reviewing the manuscript and providing valuable feedback.

\appendices

\newpage
\clearpage

\onecolumn
\section{EMR Variables}

\label{appendix:EMR_vars}
\vspace{0.5cm}
\begin{table}[h]
\caption{EMR variables (demographics, vitals and labs) in patient episode matrix. Demographics such as gender and race/ethnicity were encoded as binary variables.}

\centering
\resizebox{\textwidth}{!}{
\begin{tabular}{lllll}\hline
\multicolumn{4}{c} {Demographics and Vitals}\\ \hline
Age & Sex\_F & Sex\_M & race\_African American \\
race\_Asian/Indian/Pacific Islander  & race\_Caucasian/European Non-Hispanic & race\_Hispanic & race\_unknown \\ 
Abdominal Girth & FLACC Pain Face & Left Pupillary Response Level & Respiratory Effort Level \\
Activity Level & FLACC Pain Intensity & Level of Consciousness & Respiratory Rate \\
Bladder pressure & FLACC Pain Legs & Lip Moisture Level & Right Pupil Size After Light \\
Capillary Refill Rate & Foley Catheter Volume & Mean Arterial Pressure & Right Pupil Size Before Light \\
Central Venous Pressure & Gastrostomy Tube Volume & Motor Response Level & Right Pupillary Response Level \\
Cerebral Perfusion Pressure & Glascow Coma Score & Nasal Flaring Level & Sedation Scale Level \\
Diastolic Blood Pressure & Head Circumference & Near-Infrared Spectroscopy SO2 & Skin Turgor\_edema \\
EtCO2 & Heart Rate & Nutrition Level & Skin Turgor\_turgor \\
Extremity Temperature Level & Height & Oxygenation Index & Systolic Blood Pressure \\
Eye Response Level & Hemofiltration Fluid Output & PaO2 to FiO2 & Temperature \\
FLACC Pain Activity & Intracranial Pressure & Patient Mood Level & Verbal Response Level \\
FLACC Pain Consolability & Left Pupil Size After Light & Pulse Oximetry & WAT1 Total \\
FLACC Pain Cry & Left Pupil Size Before Light & Quality of Pain Level & Weight \\ \\ \hline
\multicolumn{4}{c} {Labs}\\ \hline
ABG Base excess & CBG PCO2 & GGT & Neutrophils \% \\
ABG FiO2 & CBG PO2 & Glucose & PT \\
ABG HCO3 & CBG TCO2 & Haptoglobin & PTT \\
ABG O2 sat & CBG pH & Hematocrit & Phosphorus level \\
ABG PCO2 & CSF Bands\% & Hemoglobin & Platelet Count \\
ABG PO2 & CSF Glucose & INR & Potassium \\
ABG TCO2 & CSF Lymphs \% & Influenza Lab & Protein Total \\
ABG pH & CSF Protein & Lactate & RBC Blood \\
ALT & CSF RBC & Lactate Dehydrogenase Blood & RDW \\
AST & CSF Segs \% & Lactic Acid Blood & Reticulocyte Count \\
Albumin Level & CSF WBC & Lipase & Schistocytes \\
Alkaline phosphatase & Calcium Ionized & Lymphocyte \% & Sodium \\
Amylase & Calcium Total & MCH & Spherocytes \\
Anti-Xa Heparin & Chloride & MCHC & T4 Free \\
B-type Natriuretic Peptide & Complement C3 Serum & MCV & TSH \\
BUN & Complement C4 Serum & MVBG Base Excess & Triglycerides \\
Bands \% & Creatinine & MVBG FiO2 & VBG Base excess \\
Basophils \% & Culture Blood & MVBG HCO3 & VBG FiO2 \\
Bicarbonate Serum & Culture CSF & MVBG O2 Sat & VBG HCO3 \\
Bilirubin Conjugated & Culture Fungus Blood & MVBG PCO2 & VBG O2 sat \\
Bilirubin Total & Culture Respiratory & MVBG PO2 & VBG PCO2 \\
Bilirubin Unconjugated & Culture Urine & MVBG TCO2 & VBG PO2 \\
Blasts \% & Culture Wound & MVBG pH & VBG TCO2 \\
C-Reactive Protein & D-dimer & Macrocytes & VBG pH \\
CBG Base excess & ESR & Magnesium Level & White Blood Cell Count \\
CBG FiO2 & Eosinophils \% & Metamyelocytes \% & \\
CBG HCO3 & Ferritin Level & Monocytes \% &  \\
CBG O2 sat & Fibrinogen & Myelocytes \% &  \\ \hline
\end{tabular}
}

\label{rnn-inputs_5col}
\end{table}
\newpage
\begin{table*}
\caption{EMR Variables (drugs and interventions) in baseline patient episode matrix.}
\centering
\resizebox{\textwidth}{!}{
\begin{tabular}{lllll}\hline
\multicolumn{4}{c} {Drugs}\\ \hline
Acetaminophen/Codeine\_inter & Clonazepam\_inter & Ipratropium Bromide\_inter & Oseltamivir\_inter \\
Acetaminophen/Hydrocodone\_inter & Clonidine HCl\_inter & Isoniazid\_inter & Oxacillin\_inter \\
Acetaminophen\_inter & Cyclophosphamide\_inter & Isradipine\_inter & Oxcarbazepine\_inter \\
Acetazolamide\_inter & Desmopressin\_inter & Ketamine\_cont & Oxycodone\_inter \\
Acyclovir\_inter & Dexamethasone\_inter & Ketamine\_inter & Pantoprazole\_inter \\
Albumin\_inter & Dexmedetomidine\_cont & Ketorolac\_inter & Penicillin G Sodium\_inter
\\
Albuterol\_inter & Diazepam\_inter & Labetalol\_inter & Pentobarbital\_inter \\
Allopurinol\_inter & Digoxin\_inter & Lactobacillus\_inter & Phenobarbital\_inter \\
Alteplase\_inter & Diphenhydramine HCl\_inter & Lansoprazole\_inter & Phenytoin\_inter
\\
Amikacin\_inter & Dobutamine\_cont & Levalbuterol\_inter & Piperacillin/Tazobactam\_inter \\
Aminophylline\_cont & Dopamine\_cont & Levetiracetam\_inter & Potassium Chloride\_inter
 \\
Aminophylline\_inter & Dornase Alfa\_inter & Levocarnitine\_inter & Potassium Phosphat
e\_inter \\
Amlodipine\_inter & Enalapril\_inter & Levofloxacin\_inter & Prednisolone\_inter \\
Amoxicillin/clavulanic acid\_inter & Enoxaparin\_inter & Levothyroxine Sodium\_inter &
 Prednisone\_inter \\
Amoxicillin\_inter & Epinephrine\_cont & Lidocaine\_inter & Propofol\_cont \\
Amphotericin B Lipid Complex\_inter & Epinephrine\_inter & Linezolid\_inter & Propofol
\_inter \\
Ampicillin/Sulbactam\_inter & Epoetin\_inter & Lisinopril\_inter & Propranolol HCl\_inter \\
Ampicillin\_inter & Erythromycin\_inter & Lorazepam\_inter & Racemic Epi\_inter \\
Aspirin\_inter & Factor VII\_inter & Magnesium Sulfate\_inter & Ranitidine\_inter \\
Atropine\_inter & Famotidine\_inter & Meropenem\_inter & Rifampin\_inter \\
Azathioprine\_inter & Fentanyl\_cont & Methadone\_inter & Risperidone\_inter \\
Azithromycin\_inter & Fentanyl\_inter & Methylprednisolone\_inter & Rocuronium\_inter \\
Baclofen\_inter & Ferrous Sulfate\_inter & Metoclopramide\_inter & Sildenafil\_inter \\
Basiliximab\_inter & Filgrastim\_inter & Metronidazole\_inter & Sodium Bicarbonate\_inter \\
Budesonide\_inter & Fluconazole\_inter & Micafungin\_inter & Sodium Chloride\_inter \\
Bumetanide\_inter & Fluticasone\_inter & Midazolam HCl\_cont & Sodium Phosphate\_inter \\
Calcium Chloride\_cont & Fosphenytoin\_inter & Midazolam HCl\_inter & Spironolactone\_inter \\
Calcium Chloride\_inter & Furosemide\_cont & Milrinone\_cont & Sucralfate\_inter \\
Calcium Gluconate\_inter & Furosemide\_inter & Montelukast Sodium\_inter & Tacrolimus\_inter \\
Carbamazepine\_inter & Gabapentin\_inter & Morphine\_cont & Terbutaline\_cont \\
Cefazolin\_inter & Ganciclovir Sodium\_inter & Morphine\_inter & Tobramycin\_inter \\
Cefepime\_inter & Gentamicin\_inter & Mycophenolate Mofetl\_inter & Topiramate\_inter \\
Cefotaxime\_inter & Glycopyrrolate\_inter & Naloxone HCL\_cont & Trimethoprim/Sulfamethoxazole\_inter \\
Cefoxitin\_inter & Heparin\_cont & Naloxone HCL\_inter & Ursodiol\_inter \\
Ceftazidime\_inter & Heparin\_inter & Nifedipine\_inter & Valganciclovir\_inter \\
Ceftriaxone\_inter & Hydrocortisone\_inter & Nitrofurantoin\_inter & Valproic Acid\_inter \\
Cephalexin\_inter & Hydromorphone\_cont & Nitroprusside\_cont & Vancomycin\_inter \\
Chloral Hydrate\_inter & Hydromorphone\_inter & Norepinephrine\_cont & Vasopressin\_cont \\
Chlorothiazide\_inter & Ibuprofen\_inter & Nystatin\_inter & Vecuronium\_inter \\
Ciprofloxacin HCL\_inter & Immune Globulin\_inter & Octreotide Acetate\_cont & Vitamin K\_inter \\
Cisatracurium\_cont & Insulin\_cont & Olanzapine\_inter & Voriconazole\_inter \\
Clindamycin\_inter & Insulin\_inter & Ondansetron\_inter & \\ \\ \hline
\multicolumn{4}{c} {Interventions}\\ \hline
Abdominal X Ray & Diversional Activity\_tv & NIV Mode & Range of Motion Assistance Type \\
Arterial Line Site & ECMO Hours & NIV Set Rate & Sedation Intervention Level \\
CT Abdomen Pelvis & EPAP & Nitric Oxide & Sedation Response Level \\
CT Brain & FiO2 & Nurse Activity Level Completed & Tidal Volume Delivered \\
CT Chest & Gastrostomy Tube Location & O2 Flow Rate & Tidal Volume Expiratory \\
Central Venous Line Site & HFOV Amplitude & Oxygen Mode Level & Tidal Volume Inspiratory \\
Chest Tube Site & HFOV Frequency & Oxygen Therapy & Tidal Volume Set \\
Chest X Ray & Hemofiltration Therapy Mode & PEEP & Tracheostomy Tube Size \\
Comfort Response Level & IPAP & Peak Inspiratory Pressure & Ventilator Rate \\
Continuous EEG Present & Inspiratory Time & Peritoneal Dialysis Type & Ventriculostomy Site \\
Diversional Activity\_books & MRI Brain & Pharmacological Comfort Measures Given & Visitor Mood Level \\
Diversional Activity\_music & Mean Airway Pressure & Position Support Given & Visitor Present \\
Diversional Activity\_play & Mechanical Ventilation Mode & Position Tolerance Level & Volume Tidal \\
Diversional Activity\_toys & MultiDisciplinaryTeam Present & Pressure Support & 
 \\ \hline
\end{tabular}
}
\label{rnn-inputs_5col_b}
\end{table*}
\setcounter{table}{0}
\vspace{0.5cm}


\twocolumn
\ifCLASSOPTIONcompsoc
\else
\fi


\ifCLASSOPTIONcaptionsoff
  \newpage
\fi



\bibliographystyle{IEEEtran}
\bibliography{Includes/prnn_arxiv.bib}
%


\end{document}